\documentclass[letterpaper, 10 pt, conference]{ieeeconf}  
\usepackage{url}
\makeatletter
\let\NAT@parse\undefined
\makeatother
\usepackage[colorlinks,citecolor=blue,urlcolor=blue,bookmarks=false,hypertexnames=true]{hyperref}
\usepackage{amsmath,amssymb,amsfonts}
\usepackage[numbers,sort&compress]{natbib}
\usepackage{booktabs}
\usepackage{multirow}
\usepackage{algorithmic}
\usepackage{graphicx}
\usepackage{graphics} 
\usepackage{epsfig} 
\usepackage{times} 
\usepackage{indentfirst}
\usepackage{bbding}
\usepackage{mathtools}
\usepackage{algorithmic}
\usepackage{ulem}

\IEEEoverridecommandlockouts                              

\usepackage{etoolbox}
\makeatletter
\patchcmd{\@makecaption}
  {\scshape}
  {}
  {}
  {}
\makeatletter
\patchcmd{\@makecaption}
  {\\}
  {.\ }
  {}
  {}
\makeatother

\title{\LARGE \bf
NanoMVG: USV-Centric Low-Power Multi-Task Visual Grounding based on Prompt-Guided Camera and 4D mmWave Radar
}

\author{Runwei Guan$^{1,2,\ \dagger}$, Jianan Liu$^{3\ \dagger}$, Liye Jia$^{1,2,}$, Haocheng Zhao$^{1,2,}$, Shanliang Yao$^{1,2,}$, \\ Xiaohui Zhu$^{2}$, Ka Lok Man$^{2}$, Eng Gee Lim$^{2}$, \textit{Senior Member, IEEE}, Jeremy Smith$^{1}$, Yutao Yue$^{4\ *}$
\thanks{
1. Department of EEE, University of Liverpool, Liverpool, UK;
}
\thanks{
2. SAT, Xi'an Jiaotong-Liverpool University, Suzhou, China;
}
\thanks{
3. Momoni AI, Gothenburg, Sweden;
}
\thanks{
4. Thrust of Artificial Intelligence and Thrust of Intelligent Transportation, HKUST (GZ), Guangzhou, China;
}
\thanks{$\dagger$ Runwei Guan and Jianan Liu contribute equally.}
\thanks{Corresponding author: \tt\small yutaoyue@hkust-gz.edu.cn}}

\begin{document}

\maketitle
\thispagestyle{empty}
\pagestyle{empty}

\begin{abstract}
Recently, visual grounding and multi-sensors setting have been incorporated into perception system for terrestrial autonomous driving systems and Unmanned Surface Vessels (USVs), yet the high complexity of modern learning-based visual grounding model using multi-sensors prevents such model to be deployed on USVs in the real-life. To this end, we design a low-power multi-task model named NanoMVG for waterway embodied perception, guiding both camera and 4D millimeter-wave radar to locate specific object(s) through natural language. NanoMVG can perform both box-level and mask-level visual grounding tasks simultaneously. Compared to other visual grounding models, NanoMVG achieves highly competitive performance on the WaterVG dataset, particularly in harsh environments. Moreover, the real-world experiments with deployment of NanoMVG on embedded edge device of USV demonstrates its fast inference speed for real-time perception and capability of boasting ultra-low power consumption for long endurance.
\end{abstract}

\section{INTRODUCTION}
\label{sec:intro}
Currently, multi-modal sensor fusion is recognized as a predominant perception method in autonomous driving \cite{wang2023multi} and intelligent transportation systems \cite{huang2023v2x}. Among these methods, the fusion of 4D millimeter-wave radar (4D radar) and cameras offers a low-cost, complementary, and robust perception approach \cite{RCFusion, xiong2023lxl, DPFT}, especially for Unmanned Surface Vessel (USV)-based waterway perception under adverse conditions where cameras may temporarily fail \cite{Efficient_vrne, guan2024mask}. The challenges of USV-based real-time waterway monitoring are compounded by the irregularity of channel zones, variable lighting conditions, and erratic vessel movement \cite{yao2023waterscenes}. These challenges often require locating specific objects based on the channel warden's intent, necessitating effective human-model interaction \cite{guan2024watervg}. For instance, channel wardens may need to identify illegal vessels or flotage by appearance or quantitative features such as distance and velocity. However, traditional image-only visual grounding can only query object(s) by the appearance and qualitative context. Hence, we introduce 4D radar, which can capture essential numeric features of objects, including depth, azimuth, velocity, radial motion, reflected power, etc \cite{yao2023radar2, SMURF, EOT_vs_POT_4D_Radar_3D_MOT,jia2025radarnext}. The fusion of radar and image surpasses the limitations of enabling queries based on both quantitative and qualitative object features. To address this, we propose a multi-sensor visual grounding approach that can locate objects using textual prompts guided by both camera and 4D radar inputs. Moreover, to mitigate the latency in information transmission and protect the privacy of perceived data, the model is designed for deployment on edge devices aboard USVs \cite{qiao2023survey}. This approach emphasizes low-power consumption to ensure prolonged USV operation for sustainable monitoring.

\begin{figure}
    \centering
    \includegraphics[width=0.99\linewidth]{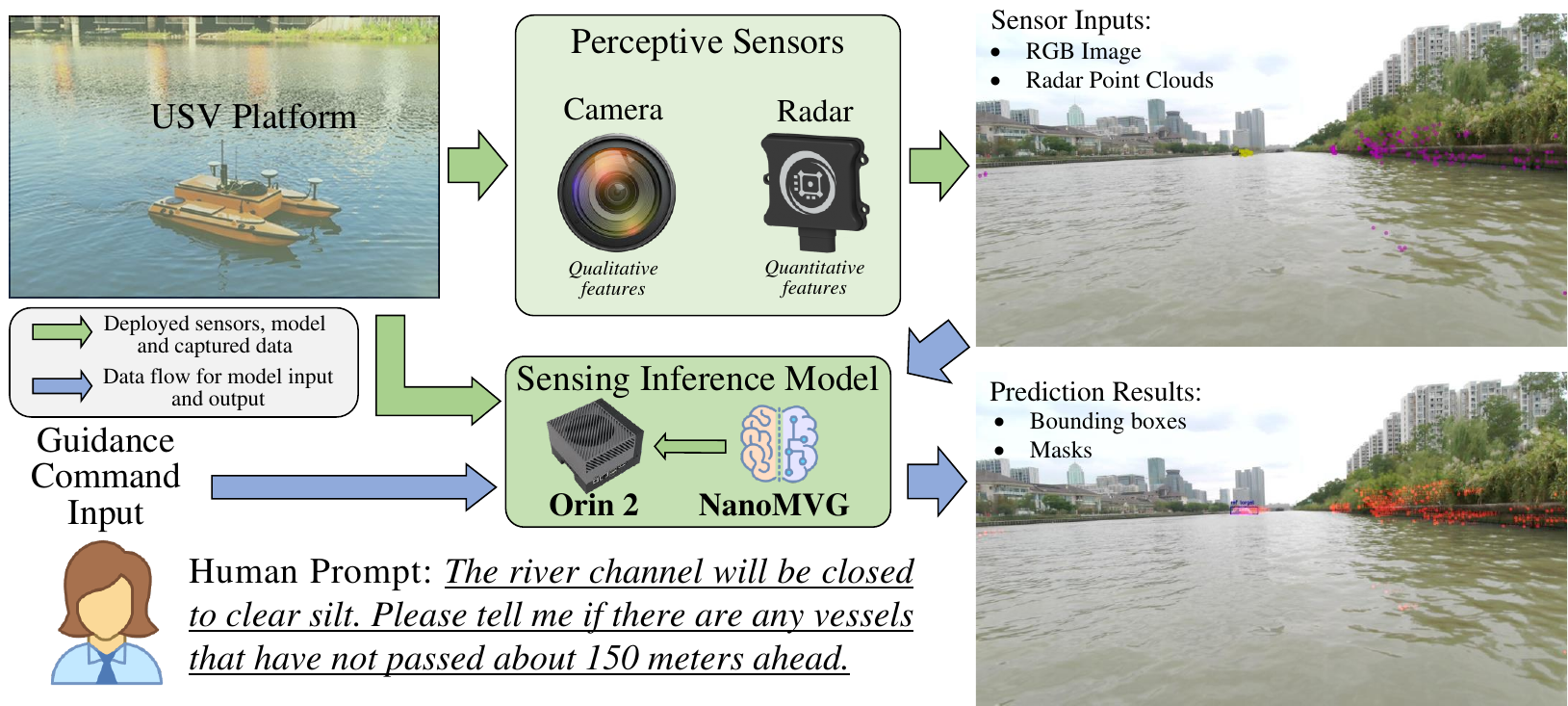}
    \vspace{-6mm}
    \caption{The pipeline of our proposed NanoMVG.}
    \label{fig:pipeline}
\end{figure}

To address the aforementioned challenges, we focus on waterway visual grounding guided by camera and radar text prompts and propose a lightweight multi-sensor visual grounding model named NanoMVG as shown in \textbf{Fig. \ref{fig:pipeline}}. NanoMVG is designed with a multi-input and multi-output architecture. It accepts three inputs: an RGB image, a 2D radar map, and a textual prompt. Correspondingly, it produces two outputs in response to the textual query, comprising the predicted object mask(s) and bounding box(es).

\begin{figure*}
    \centering
    \includegraphics[width=0.99\linewidth]{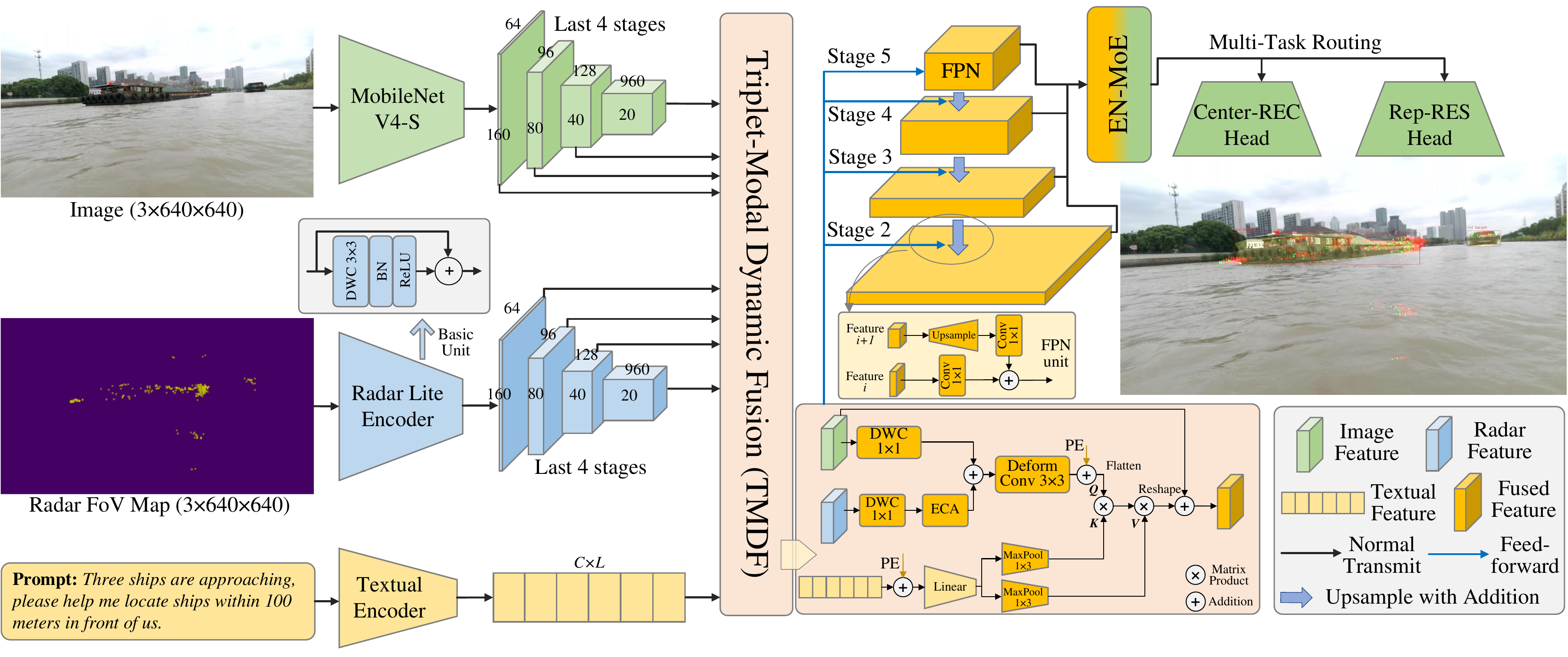}
    \vspace{-4mm}
    \caption{The architecture of NanoMVG. FoV indicates Front of View. The Triplet-Modal Dynamic Fusion (TMDF) and EN-MoE are two our proposed core modules.}
    \label{fig:model}
\end{figure*}

To mitigate the computational complexity of handling multiple branches for multi-modal fusion and multi-task prediction while maintaining model performance, \textbf{firstly}, we propose a high-performance fusion method called Triplet-Modal Dynamic Fusion (TMDF). Unlike the two-stage fusion paradigm \cite{guan2024watervg}, TMDF consolidates fusion stages for image, radar, and text features into one. It dynamically aligns and constructs text-conditional sensor features incorporating image and radar data at a low cost using simplified cross-attention. \textbf{Secondly}, to optimize sharing feature and ensure sufficient feature feeding for two tasks needing different representations, we design a lightweight Mixture-of-Expert module called Edge-Neighbor Mixture-of-Expert (EN-MoE). This module adaptively re-weights high and low-frequency features for detection and segmentation heads.

Based on the above, the contribution of this paper can be summarized as follows:

\begin{enumerate}
    \item A multi-task and low-power model named NanoMVG, designed specifically for the low-power requirement from real-time embedded edge device on USV, combines image and radar for comprehensive visual grounding. NanoMVG achieves state-of-the-art performance in balancing accuracy and power consumption.
    \item An efficient fusion module called Triplet-Modal Dynamic Fusion (TMDF), effectively integrates three modalities to achieve global and simultaneous semantic alignment and cross-modal fusion.
    \item A Mixture-of-Expert (MoE) module tailored for detection and segmentation in multi-task settings, which adaptively allocates edge and neighborhood features, significantly enhancing the performances.
\end{enumerate}

\section{Related Works}
\subsection{Waterway Perception based on Camera-Radar Fusion}
Waterway perception based on multi-sensor fusion is essential for USV-based autonomous driving \cite{clunie2021development,yao2023waterscenes,guan2023achelous}. Considering the cost and reliability, the combination of camera and radar is currently seen as a robust and affordable approach. Currently, most works focus on object detection \cite{cheng2021flow,guo2023d3,bovcon2018stereo}, segmentation \cite{vzust2023lars,taipalmaa2019high,zhou2021image}, panoptic perception \cite{guan2024mask,guan2023achelous,guan2023achelouspp}, multi-object tracking \cite{varga2022seadronessee} and Simultaneous Localization And Mapping (SLAM) \cite{cheng2021we}. Nevertheless, in numerous scenarios such as surface operations in hazardous environments, adverse weather conditions, emergency rescue, waterway monitoring, and environmental protection, USVs need to rely on human guidance through natural language for specific object perception to complete path planning. Currently, only Guan et al. \cite{guan2024watervg} have proposed a multi-sensor-based waterway visual grounding dataset and benchmark. However, this work does not deeply consider the optimization of lightweight visual grounding models under computational constraints. Therefore, we propose NanoMVG to explore low-power human-machine interactive perception on USV platforms.

\subsection{Visual Grounding Models}
Visual grounding is a multi-modal task that involves locating object(s) within an image, in response to a textual prompt. The task is categorized as Referring Expression Comprehension (REC) and Referring Expression Segmentation (RES), respectively \cite{qiao2020referring}. Specifically, REC stands for object detection with the bounding box format according to the text description and RES does segmentation with the mask format according to the text description. Model paradigms for visual grounding can be classified into three types: two-stage \cite{yang2019dynamic,yu2018mattnet,wang2019neighbourhood,hong2019learning}, one-stage \cite{luo2020multi,chen2018real,yang2019fast}, and transformer-based \cite{seqtr,transvg,cgformer,lavt,vlt}. Two-stage models operate through proposal generation followed by refinement, with text-image fusion occurring in the final stage of the first step and immediately after generating rough proposals. One-stage models integrate localization and confidence prediction within a single step, significantly reducing computational complexity. Transformer-based models utilize self-attention and cross-attention mechanisms, treating visual grounding as sequence prediction conditioned on text-image features, albeit at the cost of increased computational load due to the quadratic complexity of self-attention. To strike a balance between performance and computational efficiency, our proposed NanoMVG utilizes a one-stage architecture as its foundational framework, implementing a multi-task strategy that enhances the performance of individual tasks. Exactly, detection aids in the rapid localization of segmentation regions, while the detailed representation provided by segmentation refines the bounding box predictions in detection.

\section{Method}
\label{sec:method}

\subsection{Overview of Pipeline}
\textbf{Fig. \ref{fig:model}} illustrates the architecture of NanoMVG, which features three input branches for encoding an RGB image, a radar map projected onto the image plane—encompassing range, velocity, and reflected power as three channels, and a textual prompt for the query. Specifically, we employ MobileNetV4-S \cite{qin2024mobilenetv4}, an efficient on-device backbone, for image feature extraction. For the radar data, we utilize a cascade of $3 \times 3$ Depth-Wise Convolutions (DWC) with Batch Normalization (BN) and ReLU activation as the basic unit of the Radar Lite Encoder. This design aims to minimize parameter count while incorporating a residual connection. For the textual encoder, we use MobileBERT \cite{sun2020mobilebert}, a lightweight pretrained transformer model that provides fast and efficient language encoding. Based on four-stage image and radar features, we design a fast and computation-efficient fusion module called Triplet-Modal Dynamic Fusion (TMDF) to concurrently fuse three modalities while significantly reducing the parameter burden caused by phased fusion. In TMDF, the textual features are fused with the image and radar features at four different scales. Subsequently, the four-stage text-conditional image-radar features are fed into a Feature Pyramid Network (FPN) to enhance object representation with multi-granular features and obtain four-stage feature maps ${S_i}, i \in \{2, 3, 4, 5\}$. To address the distinct requirements of REC and RES tasks and mitigate the negative impact of mutual inhibition, we introduce the Mixture of Expert (MoE) concept and propose a feature routing module called Edge-Neighbour MoE (EN-MoE). This module preprocesses the output features from the FPN and adaptively weighs the edge and neighborhood features for the prediction heads. To further accelerate inference speed, we implement a center-based anchor-free REC head and a re-parameterized RES head to avoid redundant computational operations.

\subsection{Triplet-Modal Dynamic Fusion}
As illustrated in \textbf{Fig. \ref{fig:model}}, for a one-stage image feature \( f_I \in \mathbb{R}^{C \times H \times W} \), it first undergoes a \(1 \times 1\) Depth-Wise Convolution (DWC) (\( \mathbb{W}_I \)), while the radar feature \( f_R \in \mathbb{R}^{C \times H \times W} \) is processed by a \(1 \times 1\) DWC (\( \mathbb{W}_R \)) followed by an Efficient Channel Attention (ECA) module \cite{wang2020eca} to weigh the spatial and channel importance. The image and radar feature maps of the same size are then added together and subjected to a \(3 \times 3\) deformable convolution to finely model object features with irregular shapes. Subsequently, the combined image-radar feature map is augmented with Learnable Position Encoding ($\mathtt{LPE}$) and flattened along the spatial dimensions into a sequence-like feature \( \hat{f}_{IR} \in \mathbb{R}^{C \times N (H \times W)} \) to serve as the Query ($Q$). The whole process is shown below:
\vspace{-5pt}
\begin{align}
    \hat{f_I} &= \mathbb{W}_I f_I, \\
    \hat{f_R} &= \mathtt{ECA}(\mathbb{W}_R f_R), \\
    f_{IR} &= \hat{f_I} + \hat{f_R}, \\
    \hat{f}_{IR} &= \mathtt{Flat}(\mathtt{DeformConv}(f_{IR}) + \mathtt{LPE})  
    \label{eq:image_radar_process}
\end{align}

The textual feature $f_T \in \mathbb{R}^{C\times L}$ is augmented with Absolute Position Encoding (APE) and subsequently processed by a linear feedforward module ($\mathbb{W}_T$). To eliminate redundant information in the high-dimensional linguistic feature of the textual prompt and to reduce the computational complexity of cross-modal fusion, we apply 1D max-pooling operations (kernel size $1 \times 3$ and stride of 2) to generate the Key ($K \in \mathbb{R}^{C \times l}$) and Value ($V \in \mathbb{R}^{C \times l}$), where $l<L$. The process is detailed as follows:

\vspace{-5pt}
\begin{align}
    \hat{f_T} &= \mathbb{W}_T (f_T + \mathtt{APE}), \\
    K &= \mathtt{MaxPool}(\hat{f_T}), \\ V &= \mathtt{MaxPool}(\hat{f_T}).
    \label{eq:text_process}
\end{align}

Subsequently, we calculate the cross-modal similarity matrix ($Sim_{IR\text{-}T} \in \mathbb{R}^{N \times l}$) by computing the dot product between the image-radar feature $\hat{f}_{IR}$ and the downsampled textual feature $K$. We then multiply $Sim_{IR\text{-}T}$ with $V$, resulting in the image-like linguistic-conditional feature $f_{T\text{-}IR} \in \mathbb{R}^{C \times H \times W}$, which is based on the image-radar context after the reshape operation. The process is illustrated as follows:

\vspace{-5pt}
\begin{align}
    & Sim_{IR\text{-}T} = \frac{\hat{f}_{IR} \cdot {K}}{\sqrt{d}}, \\
    & f_{T\text{-}IR} = \mathtt{Reshape}(Sim_{IR\text{-}T} \cdot V),
    \label{eq:attention}
\end{align}
where $d$ is the dimension of $\hat{f_{IR}}$ and $K$.

\subsection{Edge-Neighour Mixture-of-Expert} 

\begin{figure}
    \centering
    \includegraphics[width=0.75\linewidth]{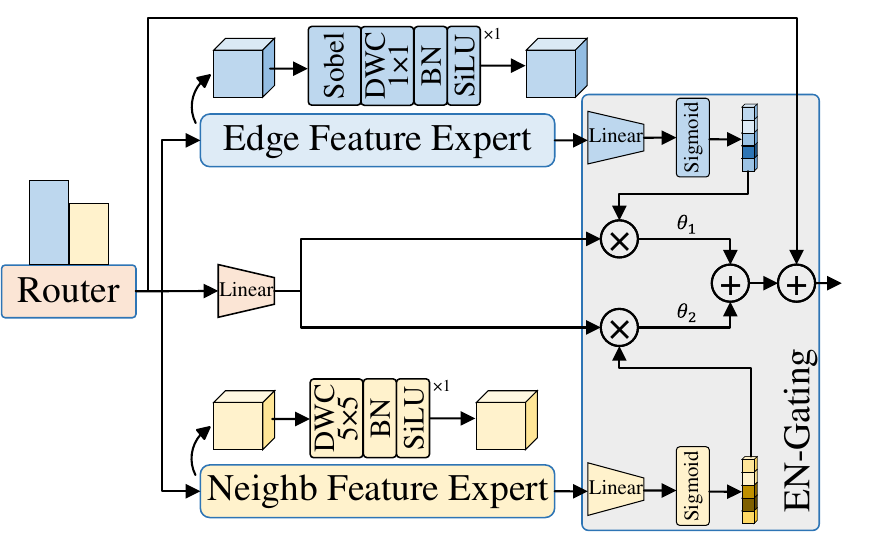}
    \vspace{-4mm}
    \caption{Edge-Neighbour Mixture-of-Expert (EN-MoE).}
    \label{fig:en_moe}
\end{figure}

Given that the feature maps at different scales within an FPN possess varying receptive fields, levels of detail, and richness of semantic information, it is significant to consider these differences. Additionally, the requirements for feature maps differ between detection and segmentation tasks. Both tasks rely on high-frequency edge features of the object, but segmentation focuses more on the internal details of the object region, whereas detection is more concerned with the contextual location of the object. Therefore, EN-MoE is designed to adaptively control the fusion and allocation of edge features and neighborhood contextual features.

As shown in Fig. \ref{fig:en_moe}, the architecture comprises two main branches: the Edge Feature Expert (EFE) and the Neighbour Feature Expert (NFE). The EFE applies the block, containing a sobel operator, a $1 \times 1$ DWC, a BN and a SiLU activation, to emphasize high-frequency feature $f_h$. The NFE captures low-frequency feature $f_l$ using a block with $5 \times 5$ DWC, BN and SiLU activation. A feedforward with a Sigmoid generates feature weighting matrices $\mathbb{W}_H$ and $\mathbb{W}_L$, which are used to weight the original feature $f_o$, producing weighted features $f_h^w$ and $f_l^w$. Controllable factors $\theta_1$ and $\theta_2$ modulate the combination of these features, resulting in the adaptive feature. Finally, a long residual path from original feature $f_o$ enhances the overall representation and obtain the final feature $f_{l\text{-}h}$. The process in EN-MoE is described as follows:
\vspace{-5pt}
\begin{align}
    & \left\{
    \begin{aligned}
        f_h &= \mathtt{SiLU}(\mathtt{BN}(\mathbb{W}_{1\times 1} \mathtt{Sobel}(f_o) )), \\
        f_l &= \mathtt{SiLU}(\mathtt{BN}(\mathbb{W}_{5\times 5}f_o)),
    \end{aligned} 
    \right.
    \\
    & \left\{
    \begin{aligned}
        \mathbb{W}_H &= \sigma(\mathbb{W}_h f_h), \\
        \mathbb{W}_L &= \sigma(\mathbb{W}_l f_l), 
    \end{aligned}
    \right.
    \\
    & \left\{
    \begin{aligned}
        \hat{f_o} &= \mathbb{W}_o f_o, \\
        f_h^w &= \mathbb{W}_H \hat{f_o}, f_l^w = \mathbb{W}_L \hat{f_o}, \\
        f_{l\text{-}h} &= (\theta_1 f_h^w + \theta_2 f_l^w) + f_o,
    \end{aligned}
    \right.
\end{align}
where $\theta_1$ and $\theta_2$ are two Sigmoid-implied learnable factors.

\begin{figure}
    \centering
    \includegraphics[width=0.99\linewidth]{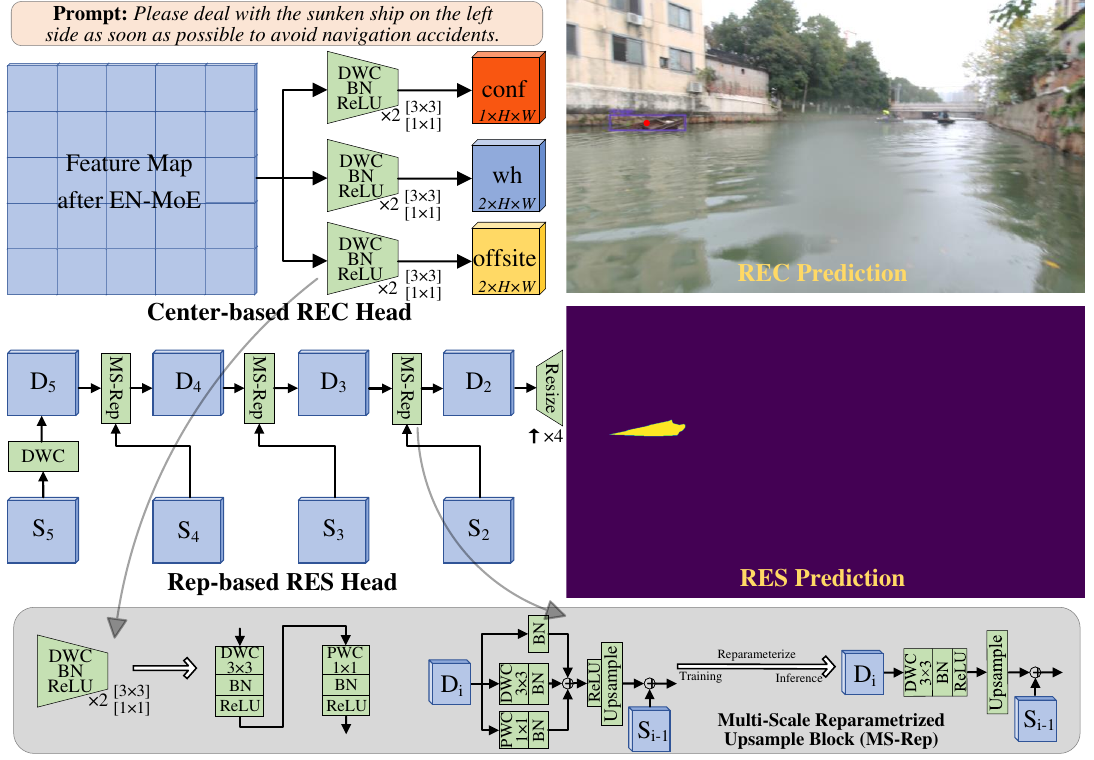}
    \vspace{-7mm}
    \caption{The structure of REC and RES heads.}
    \label{fig:pred_heads}
\end{figure}

\subsection{Prediction Heads of REC and RES}
Fig. \ref{fig:pred_heads} shows the structure of REC and RES heads.

\textbf{Center-based REC Head.} Unlike object detection, which requires locating all objects within a scene, REC only needs to locate specific object(s) based on a textual query. Therefore, we design the Center-REC Head, which is based on the center-point heatmap of the object with the anchor-free paradigm \cite{zhou2019objects}. Compared to other task heads, such as anchor-based and transformer decoder-based heads, the Center-REC Head eliminates unnecessary and cumbersome post-processing steps, resulting in lower complexity. There are three prediction branches extended from feature maps after EN-MoE by the feedforward module, where each contains the cascade of $3\times 3$ DWC with BN and ReLU, and $1 \times 1$ Point-Wise Convolution (PWC) with BN and ReLU. Three prediction heads correspond to the prediction of confidence, width and height, and offset of center-point.

\textbf{Rep-based RES Head.} Multi-scale feature extraction can capture features of varying sizes, enhancing the model's robustness in segmentation. However, multiple branches cause slower inference and increased complexity. To address this, we propose a re-parameterized full-convolution-based RES head. Exactly, we first adopt DWC $1\times 1$ on $S_5$ and obtain $D_5$. For each MS-Rep module in the training process, three parallel branches extended from $D_i$ are added together to obtain the multi-scale feature with residual connection, followed by a ReLU activation and upsample operation, then $S_{i-1}$ is added to the upsampled result and obtain the $D_{i-1}$. During the inference stage, we introduce the reparameterization mode \cite{vasu2023mobileone} to simplify the MS-Rep as the structure of single branch. Finally, we upsample the feature map while reducing the channel to obtain the binary mask of the image size.

\begin{table*}
\scriptsize
\caption{\textmd{Overall comparison of models on WaterVG dataset. \textbf{MT} means multi-task, i.e., TransVG and SimREC are REC-only single-task models while LAVT, VLT and CGFormer are RES-only single-task models. Others are multi-task models0. \textbf{IE}: Image Encoder; RN101: ResNet101; DN53: DarkNet53; Swin-B: Swin Transformer-Base; FV24: FastViT-24SA; M4-S: MobileNetV4-S.}}
\vspace{-4mm}
    \label{tab:overall_compare}
    \setlength\tabcolsep{1.2pt}
    \centering
    \begin{tabular}{c|c|c|ccc|c|ccc|c|ccc|c|ccc|c}
    \toprule
    & & \multicolumn{8}{c}{\textbf{Vision-based Visual Grounding (VVG)}} & \multicolumn{9}{c}{\textbf{Fusion-based Visual Grounding (FVG)}} \\
    \hline
    \multirow{2}[2]{*}{\textbf{Models}} & \multirow{2}[2]{*}{\textbf{MT}} & \multirow{2}[2]{*}{\textbf{IE}} & \multicolumn{4}{c}{\textbf{Val}} & \multicolumn{4}{c}{\textbf{Test}} & \multicolumn{4}{c}{\textbf{Val}} & \multicolumn{4}{c}{\textbf{Test}} \\
    \cmidrule(lr){4-19}
     & & & \textbf{AP$_{50}$} & \textbf{AP$_{50\text{-}95}$} & \textbf{AR$_{50\text{-}95}$} &  \textbf{mIoU} & \textbf{AP$_{50}$} & \textbf{AP$_{50\text{-}95}$} & \textbf{AR$_{50\text{-}95}$} & \textbf{mIoU} & \textbf{AP$_{50}$} & \textbf{AP$_{50\text{-}95}$} & \textbf{AR$_{50\text{-}95}$} &  \textbf{mIoU} & \textbf{AP$_{50}$} & \textbf{AP$_{50\text{-}95}$} & \textbf{AR$_{50\text{-}95}$} &  \textbf{mIoU} \\
    \hline
      MattNet (CVPR 2018) \cite{yu2018mattnet} & \Checkmark & RN101 & 59.6 & 31.2 & 39.7 & 54.81 & 57.2 & 30.2 & 38.9 & 54.72 & 63.9 & 38.1 & 44.0 & 57.03 & 59.1 & 32.3 & 38.9 & 60.32\\
      MCN (CVPR 2020) \cite{luo2020multi} & \Checkmark &  DN53 & 69.9 & 39.3 & 50.7 & 60.16 & 66.7 & 39.4 & 46.2 & 59.05 & 70.6 & 45.6 & 52.5 & 61.72 & 68.1 & 42.5 & 46.9 & 66.23 \\
      TransVG (ICCV 2021) \cite{transvg} & \XSolidBrush & RN101 & 70.1 & 40.7 & 51.0 & - & 67.2 & 39.8 & 48.0 & - & 71.8 & 47.1 & 54.0 & - & 69.8 & 43.6 & 52.0 & -\\
      RefTrans (NeurlPS 2021) \cite{li2021referring} &  \Checkmark & DN53 &  70.8 & 42.8 & 52.9 & 63.72 & 70.8 & 42.5 & 50.1 & 61.03 & 73.1 & 47.1 & 53.8 & 66.19 & 69.7 & 43.1 & 52.7 & 70.19 \\
      SeqTR (ECCV 2022) \cite{seqtr} & \Checkmark & DN53 & 71.2 & 42.0 & 51.8 & 62.98 & 69.1 & 41.7 & 49.2 & 60.03 & 72.6 & 47.7 & 54.2 & 64.96 & 70.0 & 43.5 & 52.6 & 68.79\\
      LAVT (CVPR 2022) \cite{lavt} & \XSolidBrush & Swin-B &  - & - & - & 65.61 &  - & - & - & 63.06 & - & - & - & \textbf{69.20} & - & - & - & 72.60 \\
      SimREC (TMM 2023) \cite{luo2022what} & \XSolidBrush & DN53 & 56.7 & 31.3 & 39.8 & - & 54.0 & 30.1 & 37.9 & -  & 60.5 & 36.8 &  42.5 & - & 56.1 & 30.9 & 39.7 & - \\
      VLT (TPAMI 2023) \cite{vlt}  & \XSolidBrush & Swin-B & - & - & - & 64.13 &  - & - & - & 61.79 & - & - & - & 67.59 & - & - & - & 70.90  \\
      CGFormer (CVPR 2023) \cite{cgformer} & \XSolidBrush & Swin-B &- & - & - & 65.99 &  - & - & - & 62.54 & - & - & - & 69.01 & - & - & - & 73.17  \\
      Potamoi-S (TITS 2025) \cite{guan2024watervg} & \Checkmark & FV24 & 71.6 & 43.3 & 54.9 & 63.16 & 70.0 & 42.6 & 50.7 & 60.36 & 72.8 & 47.5 & 55.0 & 66.27 & 70.1 & 44.8 & 53.0 & 70.81  \\
      \hline
      \textbf{NanoMVG-S (Ours)} & \Checkmark & M4-S &72.1 & 44.3 & 55.6 & 64.72 & 70.6 & 43.2 & 50.2 & 61.38 & 73.5 & 47.6 & 54.7 & 66.62 & 72.1 & 45.4 & 54.3 & 71.66 \\
      \textbf{NanoMVG-B (Ours)} & \Checkmark & Swin-B & \textbf{73.9} & \textbf{47.2} & \textbf{57.6} & \textbf{66.85} & \textbf{72.9} & \textbf{46.4} & \textbf{52.2} & \textbf{63.57} & \textbf{75.8} & \textbf{49.2} & \textbf{56.8} & 68.98 & \textbf{73.4} & \textbf{46.5} & \textbf{57.6} & \textbf{73.50}  \\
    \bottomrule
    \end{tabular}
\end{table*}

\subsection{Training Objectives}
There are two training objectives in NanoMVG, including REC and RES. For REC, we formulate the confidence loss $L_\textit{Conf}$ as the following:

\begin{align}
    & L_\textit{Conf}=\frac{-1}{N} \sum \limits _{xy}
    \left\{
    \begin{aligned}
        \footnotesize & (1-\hat{Y}_{xy})^\alpha log(\hat{Y}_{xy}), \ \textit{if}\ Y_{xy} = 1
        \\
        \footnotesize & (1-Y_{xy})^\beta (\hat{Y}_{xy})^\alpha log(1-\hat{Y}_{xy}), \ \textit{else},
        \label{eq:arw2}
    \end{aligned}
    \right.
\end{align}
where $N$ denotes the key center points in the image (in the following equations related to loss). $\alpha$ and $\beta$ are the hyper-parameters of the focal loss. $Y_{xy}$ denotes the ground truth of center point projected by the same Gaussian kernel in CenterNet \cite{zhou2019objects} while $\hat{Y}_{xy}$ indicates the prediction, which implies the probability of object on the position $(x, y)$. Additionally, the offset loss $L_\textit{Offset}$ can be formulated as:

\begin{align}
    L_\textit{Offset} = \frac{1}{N} \sum \limits _p |\hat{O}_{\tilde{p}} - (\frac{p}{R} - \tilde{p})|,
\end{align}
$\hat{O}$ denotes the prediction of network on offset. $p$ is the center point of bounding box. $R$ denotes the downsampling ratio of feature map to the original image. $\tilde{p} = \lfloor \frac{p}{R} \rfloor$, where $\frac{p}{R}-\tilde{p}$ denotes the bias. The size loss $L_{wh}$ is formulated as:

\begin{align}
    L_{wh} = \frac{1}{N} \sum \limits ^N_{k=1} \mathtt{CIoU}(\hat{S}_{pk}, s_k),
\end{align}
where we adopt CIoU \cite{zheng2021enhancing} as the loss of bounding box regression. $\hat{S}_{pk}$ is the prediction of the $k$-th object at the center point $p$. $s_k$ is the ground truth of width and height for the $k$-th object. Therefore, the loss of REC can be assembled as:
\begin{align}
    L_\textit{REC} = \tau_1 L_\textit{Conf} + \tau_2 L_\textit{Offset} + \tau_3 L_{wh},
\end{align}
where $\tau_1$, $\tau_2$ and $\tau_3$ are the pre-defined coefficients.

For RES, we integrate Dice loss \textcolor{red}{\cite{sudre2017generalised}} with Focal loss \textcolor{red}{\cite{lin2017focal}} to enhance the model's capability in addressing data imbalance and to improve the accuracy of boundary segmentation.
\vspace{-5pt}
\begin{align}
    & L_\textit{Dice} = 1- \frac{2 \sum^N_{i=1} p_i g_i}{\sum ^N_{i=1}p_i + \sum^N_{i=1} g_i}, \\
    & L_\textit{Focal} = -\alpha (1-p_t)^{\gamma}log(p_t), \\
    & L_\textit{RES} = \lambda_1 L_\textit{Dice} + \lambda_2 L_\textit{Focal}.
\end{align}
In $L_\textit{Dice}$, $p_i$ represents the predicted probability, while $g_i$ denotes the ground truth for the $i$-th pixel. The variable $N$ indicates the total number of pixels. In $L_\textit{Focal}$, $p_t$ corresponds to the predicted probability for the $t$-th class. To maximize the distinction between foreground and background, we set the number of categories to two. The parameters $\alpha$ and $\gamma$ are hyper-parameters designed to balance the ratio of positive and negative samples and to reduce the contribution of easily classified samples to the loss, respectively. Finally, $L_\textit{Dice}$ and $L_\textit{Focal}$ represent the weighting coefficients for their respective loss functions.

In all, following the uncertainty weighting of homoscedasticity, the multi-task loss function of NanoMVG can be formulated as:
\vspace{5pt}
\begin{align}
    L_\textit{Total} =  \frac{1}{2\sigma_1^2} L_\textit{REC} + \frac{1}{2\sigma_2^2} L_\textit{RES} + log\sigma_1 + log\sigma_2,
\end{align}
where $\sigma_1$ and $\sigma_2$ are two learnable stability coefficients for two training objectives while act as the regularization terms.

\section{Experiments}

\subsection{Settings of Experiments}
\textbf{Models:} We select models with various paradigms for comparison, including CNN-based two-stage (MattNet \cite{yu2018mattnet}), CNN-based one-stage (MCN \cite{luo2020multi}, SimREC \cite{luo2022what}, Potamoi \cite{guan2024watervg}), and transformer-based (TransVG \cite{transvg}, RefTrans \cite{li2021referring}, SeqTR \cite{seqtr}, LAVT \cite{lavt}, VLT \cite{vlt}, CGFormer \cite{cgformer}). For NanoMVG-S, we adopt MobileNetV4-S and MobileBERT-T as the image and text encoders while we utilize Swin-B and Bert-B as the encoders for NanoMVG-B. Here, for all models, we load the pretrained weights (ImageNet-1K) of image encoders during training. For the hyper-parameters in NanoMVG, we set the tokenizer length of MobileBERT-T \cite{sun2020mobilebert} and BERT-B \cite{devlin2018bert} as 50. Besides, the $\tau_1$. $\tau_2$ and $\tau_3$ are set to 1, 0.1 and 1, respectively. $\lambda_1$ and $\lambda_2$ are assigned as 1 and 1. For the MHSA to compare with EN-MoE, we set the number of attention heads in four stage as 2, 4, 4, 8.

\textbf{Datasets:} We firstly train and evaluate models on WaterVG \cite{guan2024watervg}, the only dataset dedicated to waterway visual grounding containing 11,568 samples, where each sample consists of an image, a frame of radar data and a textual prompt. which includes two benchmarks: Vision-based Visual Grounding (VVG) and Fusion-based Visual Grounding (FVG). VVG only adopts the image as the query object of the textual prompt while FVG puts both image and radar data to fuse with the textual prompt for visual grounding. 
Furthermore, to validate the effectiveness of NanoMVG, we also evaluate it on three well-known visual grounding benchmarks, which are RefCOCO (testA, testB), RefCOCO+ (testA, testB) and RefCOCOg (test-umd) \cite{yu2016modeling}.

\textbf{Training:} For WaterVG dataset, we resize both image and radar map as 640 $\times$ 640 (px). We train models for 80 epochs with a batch size of 16 and the initial learning rate of 1e-3. We use the cosine scheduler. We adopt the SGDM with the momentum of 0.937 and the weight decay of 5e-4. 
For RefCOCOs, we resize the image as 640 $\times$ 640 and train models for 100 epochs with the batch size of 64. We adopt the initial learning rate of 1e-4 with AdamW, whose weight decay is 1e-4 and accompanied with a cosine scheduler.

\textbf{Evaluation:} Following the benchmark in WaterVG dataset, we select AP$_{\text{50}}$, AP$_{\text{50-95}}$ and AR$_{\text{50-95}}$ to evaluate the REC task while mIoU as the metric of RES. Moreover, we introduce metric called mEPT (mean energy versus performance tradeoffs) \cite{tu2023femtodet} to evaluate the tradeoff between the accuracy and power consumption.
\vspace{0pt}
\begin{align}
\text{mEPT} &= \frac{\frac{1}{N} \sum \limits ^n_i P}{\text{Power}_\text{relative}} ,
\label{eq:mept1}
\\
    \text{Power}_\text{relative} &= \frac{1}{\tau} \sum^N_i (\delta(W, x_i) - \delta(\hat{W}, x_i)),
\label{eq:mept2}
\end{align}
where $\tau$ denotes the total number of evaluations for the $N$ samples. $\delta(W, x_i)$ represents the energy cost when the model evaluates the $i$-th sample $x_i$ using the trained parameters $W$. Additionally, $\delta(\hat{W}, x_i)$ denotes the energy cost when the model is in an untrained initialized state. Here, $P$ denotes the model's accuracy on the current task, for instance, AP$_{50}$ for REC and mIoU for RES. A higher mEPT denotes a better trade-off between accuracy and power consumption.

\begin{figure*}
    \centering
    \includegraphics[width=0.99\linewidth]{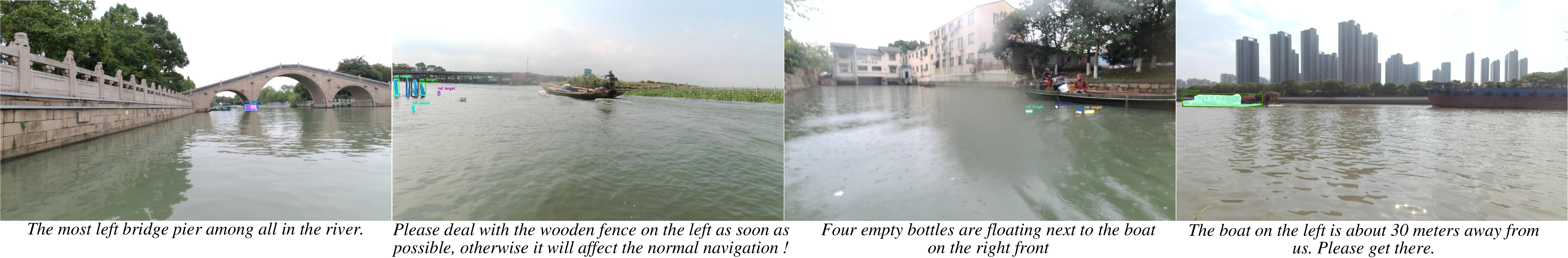}
    \vspace{-4mm}
    \caption{The prediction of NanoMVG-S. The prediction includes the predicted bounding boxes (REC) and segmented masks (RES).}
    \label{fig:vis}
\end{figure*}

\subsection{Overall Comparison with State-of-the-Arts}
TABLE \ref{tab:overall_compare} presents the performance results of various models on the WaterVG dataset, encompassing both VVG and FVG. As a multi-task model, our NanoMVG-S demonstrates competitive performance on both the REC and RES tasks; however, its performance is still outpaced by models with larger image and text encoders, such as LAVT \cite{lavt} and CGFormer \cite{cgformer}. To further validate the effectiveness of the NanoMVG architecture, we replaced the encoders in NanoMVG-S with those used in CGFormer while keeping all other components consistent. As a result, our NanoMVG-B achieves state-of-the-art performance on the WaterVG dataset across most metrics.


\subsection{Comparison on Power Consumption and Inference Speed}
TABLE \ref{tab:power_compare} presents the comparison of power consumption of models on WaterVG dataset. Our proposed NanoMVG-S achieves lower power consumption compared to all other models, whether single-task or multi-task. Notably, our model outperforms the previously introduced low-power model, Potamoi-S. This indicates that our model achieves an excellent balance between complexity, inference speed, and energy consumption. Moreover, NanoMVG-S achieves the best balance between accuracy and power consumption across all four visual grounding tasks, whether in pure vision or radar-vision fusion settings. Besides, its inference speed on Orin 2 shows a significant improvement over other models, enabling real-time inference even when the USV is traveling at high speeds \cite{cheng2021robust}. Furthermore, we observed that large models based on a fully transformer architecture exhibit significantly slower processing speeds.

\subsection{Ablation Experiments on Key Components}
We also compare our proposed components, including EN-MoE and TMDF with other methods. As TABLE \ref{tab:ablation_experiments} shows, EN-MoE demonstrates superior accuracy and a better trade-off between precision and power consumption compared to MHSA in both REC and RES tasks, while also maintaining a lower overall parameter count, particularly with significantly reduced FLOPs. In terms of tri-modal fusion, our TMDF outperforms the two-stage PHMF, despite having a larger parameter count, but still achieves relatively low FLOPs.

\subsection{Generalization Experiments}
To assess the generalization capability of our proposed NanoMVG model in visual grounding, we evaluate its performance on the RefCOCO benchmark, including RefCOCO, RefCOCO+, and RefCOCOg datasets. The results for referring expression comprehension (REC) and referring expression segmentation (RES) are presented in TABLE \ref{tab:compare_rec} and TABLE \ref{tab:compare_res}, respectively. As shown in the tables, our proposed NanoMVG-B outperforms existing visual grounding models, achieving state-of-the-art performance across both tasks. This demonstrates its superior generalization ability in visual grounding.

\begin{table}
\scriptsize
\caption{\textmd{Comparison of the trade-off between power consumption and model performance on WaterVG. }}
\vspace{-4mm}
    \label{tab:power_compare}
    \setlength\tabcolsep{0.6pt}
    \centering
    \begin{tabular}{c|cccc|cccc|c}
    \toprule
    \multirow{2}[2]{*}{\textbf{Models}} & \multicolumn{4}{c}{\textbf{VVG}} & \multicolumn{4}{c}{\textbf{FVG}} & \multirow{2}[2]{*}{\textbf{FPS} $\uparrow$} \\
    \cmidrule(lr){2-9}
     & \textbf{\text{P$_\text{R}$}} $\downarrow$ & \textbf{\text{P$_\text{O}$}} $\downarrow$ & \textbf{\text{mEPT$_\text{C}$}} $\uparrow$ & \textbf{\text{mEPT$_\text{S}$}} $\uparrow$& \textbf{\text{P$_\text{R}$}} & \textbf{\text{P$_\text{O}$}} & \textbf{\text{mEPT$_\text{C}$}} & \textbf{\text{mEPT$_\text{S}$}} \\
    \hline
      MCN \cite{luo2020multi} &  81.3 & 50.3 & 1.33 & 1.17 & 77.7 & 47.7 & 1.43 & 1.26 & 18.3\\
      SimREC \cite{luo2022what} &  94.4 & 46.8 & 1.15 & - & 88.7 & 43.5 & 1.29 & - & 26.2\\
      SeqTR \cite{seqtr} & 90.6 & 44.6 & 1.55 & 1.35 & 84.7 & 40.9 &  1.71 & 1.68 & 15.7\\
      RefTrans \cite{li2021referring} &  93.0 & 58.9 & 1.20 & 1.04 & 87.2 & 42.8 & 1.63 & 1.64 & 4.8\\
      VLT \cite{vlt}  &  136.5 & 79.6 & - & 0.78 & 130.9 & 72.1 & - & 0.98 & 1.6\\
      LAVT \cite{lavt} &  193.7 & 98.5 & - & 0.51 & 191.4 & 92.9 & - & 1.60 & 0.9  \\
      CGFormer  \cite{cgformer} &  261.3 & 122.7 & - & 0.51 & 265.2 & 117.2 & - & 0.62 & 0.6  \\
      Potamoi-S \cite{guan2024watervg} & 66.1 & 32.7 & 2.14 & 1.85& 61.5 & 30.7 & 2.28 & 2.31 & 31.2 \\
      \hline
      \textbf{NanoMVG-S} & \textbf{58.9} & \textbf{29.3} & \textbf{2.40} & \textbf{2.09} & \textbf{55.4} & \textbf{27.7}& \textbf{2.64} & \textbf{2.65}  & \textbf{34.9} \\
    \bottomrule
    \end{tabular}
    \\
    \scriptsize{\textbf{$\text{P$_\text{R}$}$} indicates the power consumption with RTX A4000 on an offline workstation. \textbf{\text{P$_\text{O}$}} denotes the power consumption with Jetson Orin 2 on the real-time embedded edge device of USV. Both values are measured by hardware API and their units are Watts per time frame (W). The mEPT are calculated upon the power consumption on Jetson Orin 2. mEPT$_\text{C}$ and mEPT$_\text{S}$ are the mEPT for REC and RES tasks.}
\end{table}


\begin{table}
    \caption{Comparison of core components in NanoMVG-S.}
    \vspace{-4mm}
    \setlength\tabcolsep{1.5pt}
    \label{tab:ablation_experiments}
    \centering
    \begin{tabular}{c|cc|ccc}
    \toprule
      \textbf{Methods} & \textbf{Param} $\downarrow$ & \textbf{FLOPs} $\downarrow$ & \textbf{AP$_{\text{50}}$} $\uparrow$ & \textbf{mIoU} $\uparrow$ & \textbf{mEPT} $\uparrow$ \\
      \hline
      \textbf{EN-MoE (ours)}  & \textbf{2.7M} & \textbf{1.82G}  & \textbf{72.1} & \textbf{71.66} & \textbf{2.65}    \\
      EN-MoE (no EFE) & 2.0M & 1.20G & 70.3 & 69.70 & 2.57 \\
      EN-MoE (no NFE) & 2.0M & 1.18G & 69.7 & 69.26 & 2.55 \\
      MHSA \cite{dosovitskiy2020image} & 3.8M & 94.65G & 67.2 & 66.02 & 1.68  \\
      
      \hline
      \textbf{TMDF (ours one-stage)}  & 6.778K & \textbf{12.98M} & \textbf{72.1} & \textbf{71.66} & \textbf{2.65} \\  
      PHMF (two-stage) \cite{guan2024watervg} &  \textbf{5.869K} & 15.20M & 71.8 & 71.07 & 2.59   \\    
      \bottomrule 
    \end{tabular} 
\end{table}

\begin{figure}
    \centering
    \includegraphics[width=0.99\linewidth]{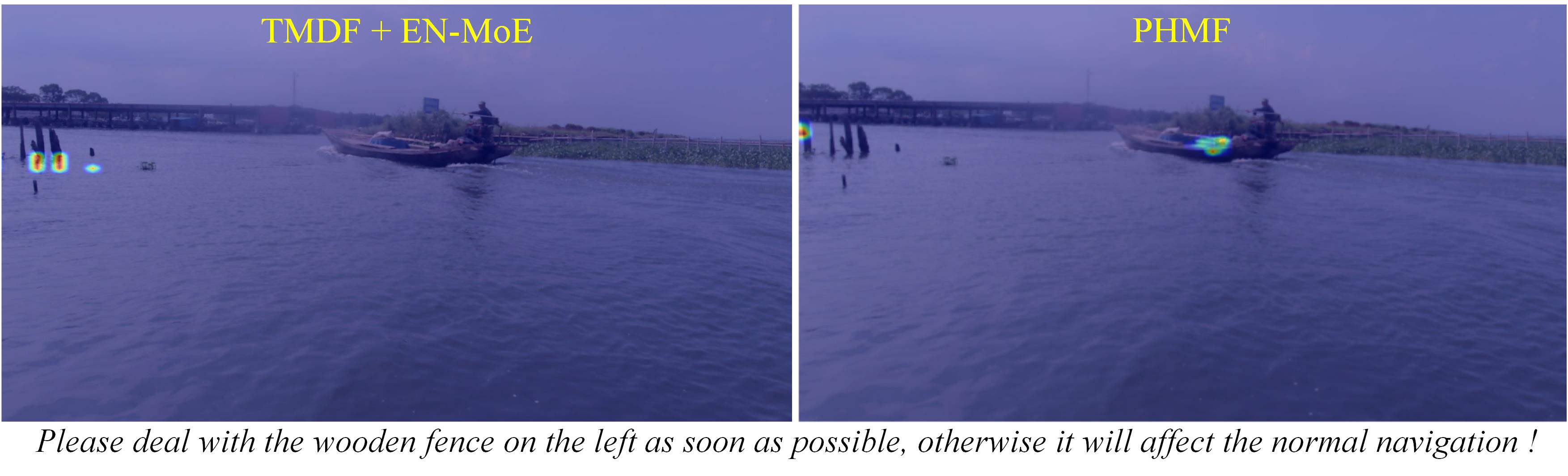}
    \vspace{-7mm}
    \caption{The comparison in heatmaps in the shared feature space before prediction heads when using two fusion paradigms .}
    \label{fig:heatmaps}
\end{figure}

\begin{table}
\footnotesize
    \caption{\textmd{Comparison of REC task on RefCOCOs datasets.}}
    \setlength\tabcolsep{6.0pt}
    \vspace{-3mm}
    \centering
    \begin{tabular}{c|cc|cc|c}
    \toprule
      \multirow{2}[2]{*}{\textbf{Models}} & \multicolumn{2}{c}{\textbf{RefCOCO}} & \multicolumn{2}{c}{\textbf{RefCOCO+}} & \textbf{RefCOCOg} \\
      \cmidrule(lr){2-3} \cmidrule(lr){4-5} \cmidrule(lr){6-6}
       & testA & testB & testA & testB & test-u \\
      \midrule
      MAttNet \cite{yu2018mattnet} & 80.43 & 69.28 & 70.26 & 56.00 & 67.27 \\
      TransVG \cite{transvg} & 82.72 & 78.35 & 70.70 & 56.94 & 67.73 \\
      RefTrans \cite{li2021referring} & 85.53 & 76.31 & 75.58 & \uline{61.91} & 69.10 \\
      \midrule
      MCN \cite{luo2020multi} & 82.29 & 74.98 & 72.86 & 57.31 & 66.01 \\
      SeqTR \cite{seqtr} & 85.00 & 76.08 & 75.37 & 58.78 & 71.58 \\
      Potamoi-S \cite{guan2024watervg} & 84.98 & 78.10 & 75.90 & 58.55 & 70.55 \\
      Potamoi-B \cite{guan2024watervg} & 87.18 & 81.65 & 78.43 & 63.30 & 74.78 \\
      \midrule
      \textbf{NanoMVG-B} & \textbf{87.53} & \textbf{82.07} & \textbf{79.12} & \textbf{64.05} & \textbf{76.40} \\
    \bottomrule
    \end{tabular}
    \label{tab:compare_rec}
\end{table}

\begin{table}
\footnotesize
    \caption{\textmd{Comparison of RES task on RefCOCOs datasets.}}
    \setlength\tabcolsep{6.0pt}
    \vspace{-3mm}
    \centering
    \begin{tabular}{c|cc|cc|c}
    \toprule
      \multirow{2}[2]{*}{\textbf{Models}} & \multicolumn{2}{c}{\textbf{RefCOCO}} & \multicolumn{2}{c}{\textbf{RefCOCO+}} & \textbf{RefCOCOg} \\
      \cmidrule(lr){2-3} \cmidrule(lr){4-5} \cmidrule(lr){6-6}
       & testA & testB & testA & testB & test-u \\
      \midrule
      MAttNet \cite{yu2018mattnet} & 62.37 & 51.70 & 52.39 & 40.08 & 48.61 \\\
      VLT \cite{vlt} & 68.29 & 62.73 & 59.20 & 49.36 & 56.65 \\
      LAVT \cite{lavt} & 73.57 & 66.98 & 66.19 & 53.59 & 57.23\\
      CGFormer \cite{cgformer} & 76.03 & 68.94 & 68.32 & 56.40 & 63.37 \\
      \midrule
      MCN \cite{luo2020multi} & 64.20 & 59.71 & 54.99 & 44.69 & 49.40 \\
      SeqTR \cite{seqtr} & 69.79 & 64.12 & 58.93 & 48.19 & 55.64 \\
      Potamoi-S \cite{guan2024watervg} & 69.05 & 63.67 & 59.05 & 48.72 & 55.23 \\
      Potamoi-B \cite{guan2024watervg} & 76.32 & 69.77 & 71.18 & 57.74 & 65.20 \\
      \midrule
      \textbf{NanoMVG-B} & \textbf{78.15} & \textbf{71.01} & \textbf{72.70} & \textbf{58.66} & \textbf{67.20} \\
    \bottomrule
    \end{tabular}
    \label{tab:compare_res}
\end{table}

\subsection{Visualization}

Fig. \ref{fig:vis} shows the prediction results of NanoMVG-S. Our NanoMVG is capable of accurately understanding and localizing the correct objects across various water environments, such as canals, lakes, and moats, based on instructions that contain visual and radar object features. Moreover, as illustrated in Fig. \ref{fig:heatmaps}, the pipeline of our TMDF and EN-MoE is capable of focusing on the correct object regions, even when the object size is relatively small. Additionally, our approach achieves a receptive field focus comparable to that of the two-stage fusion model PHMF, demonstrating that NanoMVG can attain equal or even superior semantic understanding of objects with lower computational costs.

\section{Conclusion}
This paper focuses on low-power interactive waterway perception of USVs based on edge devices, and proposes an efficient multi-task visual grounding model, NanoMVG. The model incorporates an efficient single-stage tri-modal fusion module called Triplet-Modal Dynamic Fusion (TMDF) and an adaptive multi-task feature allocation expert (EN-MoE). These components are specifically designed to address the challenges of guiding and integrating multi-sensor responses to textual queries under constrained computational conditions at a low cost, while enhancing feature representation. NanoMVG demonstrates excellent performance, achieving state-of-the-art results on the WaterVG benchmark, particularly excelling in the trade-off between model accuracy and power consumption with real-world deployment, while also enabling real-time inference.

\normalem
\footnotesize
\bibliographystyle{IEEEbib}
\bibliography{strings,main_iros_2025}

\begin{thebibliography}{10}

\bibitem{wang2023multi}
Yingjie Wang, Qiuyu Mao, Hanqi Zhu, Jiajun Deng, Yu~Zhang, Jianmin Ji, Houqiang
  Li, and Yanyong Zhang,
\newblock ``Multi-modal 3d object detection in autonomous driving: a survey,''
\newblock {\em International Journal of Computer Vision}, vol. 131, no. 8, pp.
  2122--2152, 2023.

\bibitem{huang2023v2x}
Tao Huang, Jianan Liu, Xi~Zhou, Dinh~C Nguyen, Mostafa~Rahimi Azghadi, Yuxuan
  Xia, Qing-Long Han, and Sumei Sun,
\newblock ``{V2X} cooperative perception for autonomous driving: Recent
  advances and challenges,''
\newblock {\em arXiv:2310.03525}, 2023.

\bibitem{RCFusion}
Lianqing Zheng, Sen Li, Bin Tan, Long Yang, Sihan Chen, Libo Huang, Jie Bai,
  Xichan Zhu, and Zhixiong Ma,
\newblock ``{RCFusion}: Fusing {4D} radar and camera with bird’s-eye view
  features for {3D} object detection,''
\newblock {\em IEEE Transactions on Instrumentation and Measurement}, vol. 72,
  2023,
\newblock {Art.} no. 8503814.

\bibitem{xiong2023lxl}
Weiyi Xiong, Jianan Liu, Tao Huang, Qing-Long Han, Yuxuan Xia, and Bing Zhu,
\newblock ``{LXL}: Lidar excluded lean 3d object detection with 4d imaging
  radar and camera fusion,''
\newblock {\em IEEE Transactions on Intelligent Vehicles}, vol. 9, no. 1, pp.
  79--92, 2024.

\bibitem{DPFT}
Felix Fent, Andras Palffy, and Holger Caesar,
\newblock ``{DPFT}: Dual perspective fusion transformer for camera-radar-based
  object detection,''
\newblock {\em arXiv:2404.03015}, 2024.

\bibitem{Efficient_vrne}
Runwei Guan, Shanliang Yao, Ka~Lok Man, Xiaohui Zhu, Yong Yue, Jeremy Smith,
  Eng~Gee Lim, and Yutao Yue,
\newblock ``Asy-vrnet: Waterway panoptic driving perception model based on
  asymmetric fair fusion of vision and 4d mmwave radar,''
\newblock in {\em 2024 IEEE/RSJ International Conference on Intelligent Robots
  and Systems (IROS)}. IEEE, 2024, pp. 12479--12486.

\bibitem{guan2024mask}
Runwei Guan, Shanliang Yao, Lulu Liu, Xiaohui Zhu, Ka~Lok Man, Yong Yue, Jeremy
  Smith, Eng~Gee Lim, and Yutao Yue,
\newblock ``{Mask-VRDet}: A robust riverway panoptic perception model based on
  dual graph fusion of vision and 4d mmwave radar,''
\newblock {\em Robotics and Autonomous Systems}, vol. 171, pp. 104572, 2024.

\bibitem{yao2023waterscenes}
Shanliang Yao, Runwei Guan, Zhaodong Wu, Yi~Ni, Zile Huang, Ryan~Wen Liu, Yong
  Yue, Weiping Ding, Eng~Gee Lim, Hyungjoon Seo, Ka~Lok Man, Jieming Ma,
  Xiaohui Zhu, and Yutao Yue,
\newblock ``{WaterScenes}: A multi-task 4d radar-camera fusion dataset and
  benchmarks for autonomous driving on water surfaces,''
\newblock {\em IEEE Transactions on Intelligent Transportation Systems}, pp.
  1--15, 2024,
\newblock
  doi:{\href{https://doi.org/10.1109/TITS.2024.3415772}{10.1109/TITS.2024.3415772}}.

\bibitem{guan2024watervg}
Runwei Guan, Liye Jia, Fengyufan Yang, Shanliang Yao, Erick Purwanto, Xiaohui
  Zhu, Eng~Gee Lim, Jeremy Smith, Ka~Lok Man, Xuming Hu, et~al.,
\newblock ``{WaterVG}: Waterway visual grounding based on text-guided vision
  and mmwave radar,''
\newblock {\em IEEE Transactions on Intelligent Transportation Systems}, pp.
  1--17, 2025,
\newblock
  doi:{\href{https://doi.org/10.1109/TITS.2025.3527011}{10.1109/TITS.2025.3527011}}.

\bibitem{yao2023radar2}
Shanliang Yao, Runwei Guan, Zitian Peng, Chenhang Xu, Yilu Shi, Yong Yue,
  Eng~Gee Lim, Hyungjoon Seo, Ka~Lok Man, Xiaohui Zhu, et~al.,
\newblock ``Radar perception in autonomous driving: Exploring different data
  representations,''
\newblock {\em arXiv:2312.04861}, 2023.

\bibitem{SMURF}
Jianan Liu, Qiuchi Zhao, Weiyi Xiong, Tao Huang, Qing-Long Han, and Bing Zhu,
\newblock ``{SMURF}: Spatial multi-representation fusion for 3d object
  detection with 4d imaging radar,''
\newblock {\em IEEE Transactions on Intelligent Vehicles}, vol. 9, no. 1, pp.
  799--812, 2024.

\bibitem{EOT_vs_POT_4D_Radar_3D_MOT}
Jianan Liu, Guanhua Ding, Yuxuan Xia, Jinping Sun, Tao Huang, Lihua Xie, and
  Bing Zhu,
\newblock ``Which framework is suitable for online 3d multi-object tracking for
  autonomous driving with automotive 4d imaging radar?,''
\newblock in {\em Proceedings of the IEEE Intelligent Vehicles Symposium (IV)},
  2024, pp. 1258--1265.

\bibitem{jia2025radarnext}
Liye Jia, Runwei Guan, Haocheng Zhao, Qiuchi Zhao, Ka~Lok Man, Jeremy Smith,
  Limin Yu, and Yutao Yue,
\newblock ``Radarnext: Real-time and reliable 3d object detector based on 4d
  mmwave imaging radar,''
\newblock {\em arXiv preprint arXiv:2501.02314}, 2025.

\bibitem{qiao2023survey}
Yuanyuan Qiao, Jiaxin Yin, Wei Wang, F{\'a}bio Duarte, Jie Yang, and Carlo
  Ratti,
\newblock ``Survey of deep learning for autonomous surface vehicles in marine
  environments,''
\newblock {\em IEEE Transactions on Intelligent Transportation Systems}, vol.
  24, no. 4, pp. 3678--3701, 2023.

\bibitem{clunie2021development}
Thomas Clunie, Michael DeFilippo, Michael Sacarny, and Paul Robinette,
\newblock ``Development of a perception system for an autonomous surface
  vehicle using monocular camera, lidar, and marine radar,''
\newblock in {\em IEEE International Conference on Robotics and Automation
  (ICRA)}. IEEE, 2021, pp. 14112--14119.

\bibitem{guan2023achelous}
Runwei Guan, Shanliang Yao, Xiaohui Zhu, Ka~Lok Man, Eng~Gee Lim, Jeremy Smith,
  Yong Yue, and Yutao Yue,
\newblock ``Achelous: A fast unified water-surface panoptic perception
  framework based on fusion of monocular camera and 4d mmwave radar,''
\newblock in {\em Proceedings of the IEEE 26th International Conference on
  Intelligent Transportation Systems (ITSC)}, 2023, pp. 182--188.

\bibitem{cheng2021flow}
Yuwei Cheng, Jiannan Zhu, Mengxin Jiang, Jie Fu, Changsong Pang, Peidong Wang,
  Kris Sankaran, Olawale Onabola, Yimin Liu, Dianbo Liu, et~al.,
\newblock ``Flow: A dataset and benchmark for floating waste detection in
  inland waters,''
\newblock in {\em Proceedings of the IEEE/CVF International Conference on
  Computer Vision}, 2021, pp. 10953--10962.

\bibitem{guo2023d3}
Jundong Guo, Hui Feng, Haixiang Xu, Wenzhao Yu, and Sam shuzhi Ge,
\newblock ``{D3-Net}: Integrated multi-task convolutional neural network for
  water surface deblurring, dehazing and object detection,''
\newblock {\em Engineering Applications of Artificial Intelligence}, vol. 117,
  pp. 105558, 2023.

\bibitem{bovcon2018stereo}
Borja Bovcon et~al.,
\newblock ``Stereo obstacle detection for unmanned surface vehicles by
  imu-assisted semantic segmentation,''
\newblock {\em Robotics and Autonomous Systems}, vol. 104, pp. 1--13, 2018.

\bibitem{vzust2023lars}
Lojze {\v{Z}}ust, Janez Per{\v{s}}, and Matej Kristan,
\newblock ``Lars: A diverse panoptic maritime obstacle detection dataset and
  benchmark,''
\newblock in {\em Proceedings of the IEEE/CVF International Conference on
  Computer Vision}, 2023, pp. 20304--20314.

\bibitem{taipalmaa2019high}
Jussi Taipalmaa, Nikolaos Passalis, Honglei Zhang, Moncef Gabbouj, and Jenni
  Raitoharju,
\newblock ``High-resolution water segmentation for autonomous unmanned surface
  vehicles: A novel dataset and evaluation,''
\newblock in {\em 2019 IEEE 29th International Workshop on Machine Learning for
  Signal Processing (MLSP)}. IEEE, 2019, pp. 1--6.

\bibitem{zhou2021image}
Zhiguo Zhou, Jiaen Sun, Jiabao Yu, Kaiyuan Liu, Junwei Duan, Long Chen, and
  CL~Philip Chen,
\newblock ``An image-based benchmark dataset and a novel object detector for
  water surface object detection,''
\newblock {\em Frontiers in Neurorobotics}, vol. 15, pp. 723336, 2021.

\bibitem{guan2023achelouspp}
Runwei Guan, Haocheng Zhao, Shanliang Yao, Ka~Lok Man, Xiaohui Zhu, Limin Yu,
  Yong Yue, Jeremy Smith, Eng~Gee Lim, Weiping Ding, et~al.,
\newblock ``Achelous++: Power-oriented water-surface panoptic perception
  framework on edge devices based on vision-radar fusion and pruning of
  heterogeneous modalities,''
\newblock {\em arXiv preprint arXiv:2312.08851}, 2023.

\bibitem{varga2022seadronessee}
Leon~Amadeus Varga, Benjamin Kiefer, Martin Messmer, and Andreas Zell,
\newblock ``Seadronessee: A maritime benchmark for detecting humans in open
  water,''
\newblock in {\em Proceedings of the IEEE/CVF winter conference on applications
  of computer vision}, 2022, pp. 2260--2270.

\bibitem{cheng2021we}
Yuwei Cheng, Mengxin Jiang, Jiannan Zhu, and Yimin Liu,
\newblock ``Are we ready for unmanned surface vehicles in inland waterways? the
  usvinland multisensor dataset and benchmark,''
\newblock {\em IEEE Robotics and Automation Letters}, vol. 6, no. 2, pp.
  3964--3970, 2021.

\bibitem{qiao2020referring}
Yanyuan Qiao, Chaorui Deng, and Qi~Wu,
\newblock ``Referring expression comprehension: A survey of methods and
  datasets,''
\newblock {\em IEEE Transactions on Multimedia}, vol. 23, pp. 4426--4440, 2020.

\bibitem{yang2019dynamic}
Sibei Yang, Guanbin Li, and Yizhou Yu,
\newblock ``Dynamic graph attention for referring expression comprehension,''
\newblock in {\em Proceedings of the IEEE/CVF International Conference on
  Computer Vision}, 2019, pp. 4644--4653.

\bibitem{yu2018mattnet}
Licheng Yu et~al.,
\newblock ``{MAttNet}: Modular attention network for referring expression
  comprehension,''
\newblock in {\em Proceedings of the IEEE/CVF Conference on Computer Vision and
  Pattern Recognition}, 2018, pp. 1307--1315.

\bibitem{wang2019neighbourhood}
Peng Wang, Qi~Wu, Jiewei Cao, Chunhua Shen, Lianli Gao, and Anton van~den
  Hengel,
\newblock ``Neighbourhood watch: Referring expression comprehension via
  language-guided graph attention networks,''
\newblock in {\em Proceedings of the IEEE/CVF Conference on Computer Vision and
  Pattern Recognition}, 2019, pp. 1960--1968.

\bibitem{hong2019learning}
Richang Hong, Daqing Liu, Xiaoyu Mo, Xiangnan He, and Hanwang Zhang,
\newblock ``Learning to compose and reason with language tree structures for
  visual grounding,''
\newblock {\em IEEE transactions on pattern analysis and machine intelligence},
  vol. 44, no. 2, pp. 684--696, 2019.

\bibitem{luo2020multi}
Gen Luo, Yiyi Zhou, Xiaoshuai Sun, Liujuan Cao, Chenglin Wu, Cheng Deng, and
  Rongrong Ji,
\newblock ``Multi-task collaborative network for joint referring expression
  comprehension and segmentation,''
\newblock in {\em Proceedings of the IEEE/CVF Conference on computer vision and
  pattern recognition}, 2020, pp. 10034--10043.

\bibitem{chen2018real}
Xinpeng Chen, Lin Ma, Jingyuan Chen, Zequn Jie, Wei Liu, and Jiebo Luo,
\newblock ``Real-time referring expression comprehension by single-stage
  grounding network,''
\newblock {\em arXiv:1812.03426}, 2018.

\bibitem{yang2019fast}
Zhengyuan Yang, Boqing Gong, Liwei Wang, Wenbing Huang, Dong Yu, and Jiebo Luo,
\newblock ``A fast and accurate one-stage approach to visual grounding,''
\newblock in {\em Proceedings of the IEEE/CVF International Conference on
  Computer Vision}, 2019, pp. 4683--4693.

\bibitem{seqtr}
Chaoyang Zhu, Yiyi Zhou, Yunhang Shen, Gen Luo, Xingjia Pan, Mingbao Lin, Chao
  Chen, Liujuan Cao, Xiaoshuai Sun, and Rongrong Ji,
\newblock ``{SeqTR}: A simple yet universal network for visual grounding,''
\newblock in {\em Proceedings of the European Conference on Computer Vision},
  2022, pp. 598--615.

\bibitem{transvg}
Jiajun Deng, Zhengyuan Yang, Tianlang Chen, Wengang Zhou, and Houqiang Li,
\newblock ``{TransVG}: End-to-end visual grounding with transformers,''
\newblock in {\em Proceedings of the IEEE/CVF International Conference on
  Computer Vision}, 2021, pp. 1769--1779.

\bibitem{cgformer}
Jiajin Tang, Ge~Zheng, Cheng Shi, and Sibei Yang,
\newblock ``Contrastive grouping with transformer for referring image
  segmentation,''
\newblock in {\em Proceedings of the IEEE/CVF Conference on Computer Vision and
  Pattern Recognition}, 2023, pp. 23570--23580.

\bibitem{lavt}
Zhao Yang, Jiaqi Wang, Yansong Tang, Kai Chen, Hengshuang Zhao, and Philip~HS
  Torr,
\newblock ``{LAVT}: Language-aware vision transformer for referring image
  segmentation,''
\newblock in {\em Proceedings of the IEEE/CVF Conference on Computer Vision and
  Pattern Recognition}, 2022, pp. 18155--18165.

\bibitem{vlt}
Henghui Ding, Chang Liu, Suchen Wang, and Xudong Jiang,
\newblock ``{VLT}: Vision-language transformer and query generation for
  referring segmentation,''
\newblock {\em IEEE Transactions on Pattern Analysis and Machine Intelligence},
  vol. 45, no. 6, pp. 7900--7916, 2023.

\bibitem{qin2024mobilenetv4}
Danfeng Qin, Chas Leichner, Manolis Delakis, Marco Fornoni, Shixin Luo, Fan
  Yang, Weijun Wang, Colby Banbury, Chengxi Ye, Berkin Akin, et~al.,
\newblock ``Mobilenetv4-universal models for the mobile ecosystem,''
\newblock {\em arXiv:2404.10518}, 2024.

\bibitem{sun2020mobilebert}
Zhiqing Sun, Hongkun Yu, Xiaodan Song, Renjie Liu, Yiming Yang, and Denny Zhou,
\newblock ``{MobileBERT}: a compact task-agnostic bert for resource-limited
  devices,''
\newblock in {\em Proceedings of the 58th Annual Meeting of the Association for
  Computational Linguistics}, 2020, pp. 2158--2170.

\bibitem{wang2020eca}
Qilong Wang, Banggu Wu, Pengfei Zhu, Peihua Li, Wangmeng Zuo, and Qinghua Hu,
\newblock ``Eca-net: Efficient channel attention for deep convolutional neural
  networks,''
\newblock in {\em Proceedings of the IEEE/CVF conference on computer vision and
  pattern recognition}, 2020, pp. 11534--11542.

\bibitem{zhou2019objects}
Xingyi Zhou, Dequan Wang, and Philipp Kr{\"a}henb{\"u}hl,
\newblock ``Objects as points,''
\newblock {\em arXiv:1904.07850}, 2019.

\bibitem{vasu2023mobileone}
Pavan Kumar~Anasosalu Vasu, James Gabriel, Jeff Zhu, Oncel Tuzel, and Anurag
  Ranjan,
\newblock ``Mobileone: An improved one millisecond mobile backbone,''
\newblock in {\em Proceedings of the IEEE/CVF conference on computer vision and
  pattern recognition}, 2023, pp. 7907--7917.

\bibitem{li2021referring}
Muchen Li and Leonid Sigal,
\newblock ``Referring transformer: A one-step approach to multi-task visual
  grounding,''
\newblock {\em Advances in neural information processing systems}, vol. 34, pp.
  19652--19664, 2021.

\bibitem{luo2022what}
Gen Luo, Yiyi Zhou, Jiamu Sun, Xiaoshuai Sun, and Rongrong Ji,
\newblock ``A survivor in the era of large-scale pretraining: An empirical
  study of one-stage referring expression comprehension,''
\newblock {\em IEEE Transactions on Multimedia}, vol. 26, pp. 3689--3700, 2024.

\bibitem{zheng2021enhancing}
Zhaohui Zheng, Ping Wang, Dongwei Ren, Wei Liu, Rongguang Ye, Qinghua Hu, and
  Wangmeng Zuo,
\newblock ``Enhancing geometric factors in model learning and inference for
  object detection and instance segmentation,''
\newblock {\em IEEE transactions on cybernetics}, vol. 52, no. 8, pp.
  8574--8586, 2021.

\bibitem{sudre2017generalised}
Carole~H Sudre, Wenqi Li, Tom Vercauteren, Sebastien Ourselin, and
  M~Jorge~Cardoso,
\newblock ``Generalised dice overlap as a deep learning loss function for
  highly unbalanced segmentations,''
\newblock in {\em Deep Learning in Medical Image Analysis and Multimodal
  Learning for Clinical Decision Support: Third International Workshop, DLMIA
  2017, and 7th International Workshop, ML-CDS 2017, Held in Conjunction with
  MICCAI 2017, Qu{\'e}bec City, QC, Canada, September 14, Proceedings 3}.
  Springer, 2017, pp. 240--248.

\bibitem{lin2017focal}
Tsung-Yi Lin, Priya Goyal, Ross Girshick, Kaiming He, and Piotr Doll{\'a}r,
\newblock ``Focal loss for dense object detection,''
\newblock in {\em Proceedings of the IEEE international conference on computer
  vision}, 2017, pp. 2980--2988.

\bibitem{devlin2018bert}
Jacob Devlin,
\newblock ``{BERT}: Pre-training of deep bidirectional transformers for
  language understanding,''
\newblock {\em arXiv:1810.04805}, 2018.

\bibitem{yu2016modeling}
Licheng Yu, Patrick Poirson, Shan Yang, Alexander~C Berg, and Tamara~L Berg,
\newblock ``Modeling context in referring expressions,''
\newblock in {\em Computer Vision--ECCV 2016: 14th European Conference,
  Amsterdam, The Netherlands, October 11-14, 2016, Proceedings, Part II 14}.
  Springer, 2016, pp. 69--85.

\bibitem{tu2023femtodet}
Peng Tu, Xu~Xie, Guo Ai, Yuexiang Li, Yawen Huang, and Yefeng Zheng,
\newblock ``{FemtoDet}: An object detection baseline for energy versus
  performance tradeoffs,''
\newblock in {\em Proceedings of the IEEE/CVF International Conference on
  Computer Vision}, 2023, pp. 13318--13327.

\bibitem{cheng2021robust}
Yuwei Cheng, Hu~Xu, and Yimin Liu,
\newblock ``Robust small object detection on the water surface through fusion
  of camera and millimeter wave radar,''
\newblock in {\em Proceedings of the IEEE/CVF international conference on
  computer vision}, 2021, pp. 15263--15272.

\bibitem{dosovitskiy2020image}
Alexey Dosovitskiy, Lucas Beyer, Alexander Kolesnikov, Dirk Weissenborn,
  Xiaohua Zhai, Thomas Unterthiner, Mostafa Dehghani, Matthias Minderer, Georg
  Heigold, Sylvain Gelly, et~al.,
\newblock ``An image is worth 16x16 words: Transformers for image recognition
  at scale,''
\newblock in {\em International Conference on Learning Representations}, 2020.

\end{thebibliography}

\end{document}